\documentclass[11pt,letterpaper]{article}
\usepackage{emnlp2017}
\usepackage{times}
\usepackage{latexsym}
\usepackage{url}
\usepackage{amssymb}
\usepackage{amsmath}
\usepackage{multirow,bigdelim}
\usepackage{booktabs}
\usepackage{pifont}
\usepackage{xspace}
\usepackage{graphicx}
\usepackage{commath}
\usepackage{amsfonts}
\usepackage{array}
\usepackage{footmisc}

\definecolor{tablegray}{gray}{0.8}
\definecolor{tablegray}{gray}{1}
\definecolor{examplegray}{gray}{0}

\newcommand{\ced}{\mbox{\emph{CED}}}

\emnlpfinalcopy 
\def\emnlppaperid{***} 

\title{Dynamic Data Selection for Neural Machine Translation}

 \author{Marlies van der Wees \\
 Informatics Institute \\
  University of Amsterdam \\
  \\\And
  Arianna Bisazza\thanks{\,\,\,Work done while at University of Amsterdam}\, \\
  LIACS \\
  Leiden University \\
  \\\And
  Christof Monz \\
   Informatics Institute \\
  University of Amsterdam \\
 \\}
 
\date{}

\begin{document}

\maketitle

\begin{abstract}
Intelligent selection of training data has proven a successful technique to simultaneously increase training efficiency and translation performance for phrase-based machine translation (PBMT). With the recent increase in popularity of neural machine translation (NMT), we explore in this paper \emph{to what extent} and \emph{how} NMT can also benefit from data selection.  
While state-of-the-art data selection \cite{axelrod2011domain} consistently performs well for PBMT, we show that gains are substantially lower for NMT. 
Next, we introduce \emph{dynamic data selection} for NMT, a method in which we vary the selected subset of training data between different training epochs. 
Our experiments show that the best results are achieved when applying a technique we call \emph{gradual fine-tuning}, with improvements up to +2.6 BLEU over the original data selection approach and up to +3.1 BLEU over a general baseline. 
\end{abstract}

\section{Introduction}
\label{sec:introduction}
Recent years have shown a rapid shift from phrase-based (PBMT) to neural machine translation (NMT) \cite{sutskever2014sequence,cho2014learning,bahdanau2014neural} as the most common machine translation paradigm. With large quantities of parallel data, NMT outperforms PBMT for an increasing number of language pairs \cite{bojar2016findings}. 
Unfortunately, training an NMT model is often a time-consuming task, with training times of several weeks not being unusual. 

Despite its training inefficiency, most work in NMT greedily uses all available training data for a given language pair. However, it is unlikely that all data is equally helpful to create the best-performing system. In PBMT, this issue has been addressed by applying intelligent data selection, and it has consistently been shown that using more data does not always improve translation quality \cite{moore2010intelligent,axelrod2011domain,gasco2012does}. Instead, for a given translation task, the training bitext likely contains sentences that are irrelevant or even harmful, making it beneficial to keep only the most relevant subset of the data while discarding the rest, with the additional benefit of smaller models and faster training. 

Motivated by the success of data selection in PBMT, we investigate in this paper \emph{to what extent} and \emph{how} NMT can benefit from data selection as well. 
While data selection has been applied to NMT to reduce the size of the data \cite{cho2014learning,luong2015addressing}, the effects on translation quality have not been investigated.
Intuitively, and confirmed by our exploratory experiments in Section~\ref{sec:static-results}, this is a challenging task; NMT systems are known to under-perform when trained on limited parallel data \cite{zoph2016transfer,fadaee2017data}, and do not have a separate large-scale target-side language model to compensate for smaller parallel training data. 

To alleviate the negative effect of small training data on NMT, we introduce \emph{dynamic data selection}. Following conventional data selection, we still dramatically reduce the training data size, favoring parts of the data which are most relevant to the translation task at hand. However, we exploit the fact that the NMT training process iterates over the training corpus in multiple epochs, and we alter the quantity or the composition of the training data \emph{between epochs}. 
The proposed method requires no modifications to the NMT architecture or parameters, and substantially speeds up training times while simultaneously improving translation quality with respect to a complete-bitext baseline. 

In summary, our contributions are as follows: 

(i) We compare the effects of a commonly used data selection approach \cite{axelrod2011domain} on PBMT and NMT using four different test sets. We find that this method is much less effective for NMT than for PBMT, while using the exact same training data subsets.

(ii) We introduce dynamic data selection as a way to make data selection profitable for NMT. We explore two techniques to alter the selected data subsets, and find that our method called \emph{gradual fine-tuning} 
improves over conventional static data selection (up to +2.6 BLEU) and over a high-resource general baseline (up to +3.1 BLEU). 
Moreover, gradual fine-tuning approximates in-domain fine-tuning in $\sim$20\% of the training time, even when no parallel in-domain data is available.

\section{Static data selection}
\label{sec:static-method}
As a first step towards dynamic data selection for NMT, we compare the effects of a commonly used, state-of-the-art data selection method \cite{axelrod2011domain} on both neural and phrase-based MT. 
Briefly, this approach ranks sentence pairs in a large training bitext according to their difference in cross-entropy with respect to an in-domain corpus (i.e., a corpus representing the test data) and a general corpus.
Next, the top $n$ sentence pairs with the highest rank---thus lowest cross-entropy---are selected and used for training an MT system. 

Formally, given an in-domain corpus $I$, we first create language models from the source side $f$ of $I$ (\emph{LM}$_{I,f}$) and the target side $e$ of $I$ (\emph{LM}$_{I,e}$). We then draw a random sample (similar in size to $I$) of the large general corpus $G$ and create language models from the source and target sides of $G$: \emph{LM}$_{G,f}$ and \emph{LM}$_{G,e}$, respectively. Note that the data for creating these LMs need not be parallel but can be independent corpora in both languages.

Next, we compute for each sentence pair $s$ in $G$ four cross-entropy scores, defined as:
\begin{equation}
H_{C,s_b} = -\sum p\left(s_b\right) \log \left(\mbox{\emph{LM}}_{C,b}\left(s_b\right)\right),
\end{equation}
where $C\in\{I,G\}$ is the corpus, $b\in\{f,e\}$ refers to the bitext side, and $s_b$ is the bitext side $b$ of sentence pair $s$ in the parallel training corpus.

To find sentences that are similar to the in-domain corpus, i.e., have low $H_I$, and at the same time dissimilar to the general corpus, i.e., have high $H_G$, we compute for each sentence pair $s$ the bilingual cross-entropy difference $\ced_s$ following \newcite{axelrod2011domain}:
\begin{equation}
\label{eq:ced}
\ced_s = (H_{I,s_f} - H_{G,s_f}) + (H_{I,s_e} - H_{G,s_e}).
\end{equation}
Finally, we rank all sentence pairs $s \in G$ according to their $\ced_s$, and then select only the top $n$ sentence pairs with the lowest $\ced_s$.

Following related work by \newcite{moore2010intelligent}, we restrict the vocabulary of the LMs to the words occurring at least twice in the in-domain corpus. 
To analyze the quality of the selected data subsets, we also run experiments on random selections, all performed in threefold. 
Finally, we always use the exact same selection of sentence pairs in equivalent PBMT and NMT experiments.

\paragraph{LSTM versus n-gram} 
The described data selection method uses n-gram LMs to determine the domain-relevance of sentence pairs. We adhere to this setting for our comparative experiments on PBMT and NMT (Section~\ref{sec:static-results}). However, when applying data selection to NMT, we examine the potential benefit of replacing the conventional n-gram LMs with LSTMs\footnote{We use four-layer LSTMs with embedding and hidden sizes of 1,024, which we train for 30 epochs.}. These have the advantage to remember longer histories, and do not have to back off to shorter histories when encountering out-of-vocabulary words. In this neural variant to rank sentences, the score for each sentence pair in $G$ is still computed as the bilingual cross-entropy difference in Equation~\eqref{eq:ced}. In addition, we use the same in-domain and general corpora as with the n-gram method, and we again restrict the vocabulary to the most frequent  words.

\begin{figure*}[ht!]
\includegraphics[width=\textwidth]{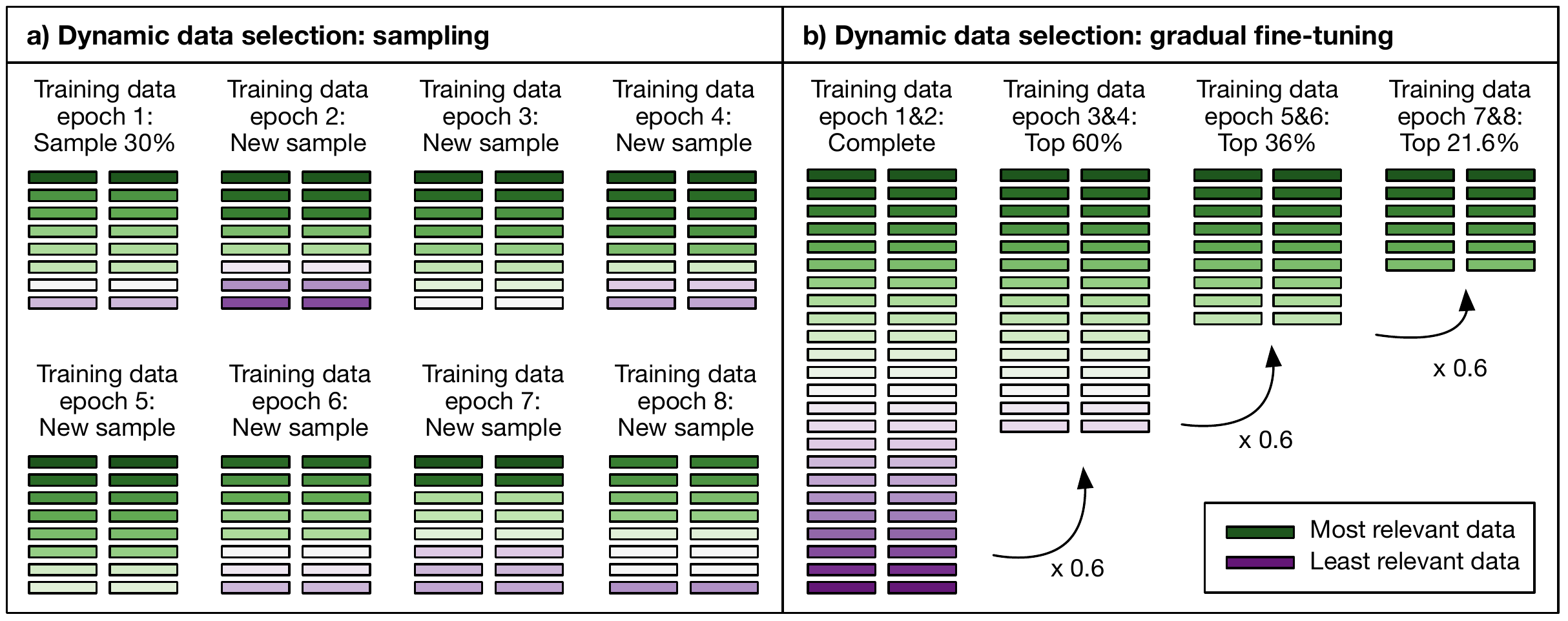}
\caption{Illustration of two dynamic bitext selection techniques for NMT: \emph{sampling} (left) and \emph{gradual fine-tuning} (right). Measured over 16 training epochs (which is used in this work), the total training time of both examples would be $\sim$30\% of the training time needed when using the complete bitext.}
\label{fig:dynamic-ds}
\end{figure*}

\section{Dynamic data selection}
\label{sec:dynamic-method}
While data selection aims to discard irrelevant data, it can also exacerbate the problem of low vocabulary coverage and unreliable statistics for rarer words in the `long tail',  which are major issues in NMT \cite{luong2015addressing,sennrich2016neural}. 
In addition, it has been shown that NMT performance drops tremendously in low-resource scenarios \cite{zoph2016transfer,fadaee2017data,koehn2017six}. 

To overcome this problem, we introduce \emph{dynamic data selection}, in which we vary the selected data subsets \textit{during} training. Unlike other MT paradigms, which require training data to be fixed during the entire training process, NMT iterates over the training corpus in several epochs, allowing to use a different subset of the training data in every epoch. 

Dynamic data selection starts from a relevance-ranked bitext, which we create using CED scores as computed in Equation~\eqref{eq:ced}. Given this ranking, we investigate two dynamic data selection techniques\footnote{Code for bitext ranking and both selection techniques: \url{github.com/marliesvanderwees/dds-nmt}.}  that vary per epoch the composition or the size of the selected training data. Both techniques aim to favor highly relevant sentences over less relevant sentences while not completely discarding the latter. In all experiments, we use a fixed vocabulary created from the complete bitext.

While we use in this work a domain-relevance ranking of the bitext following \newcite{axelrod2011domain}, dynamic data selection can also be applied using other ranking criteria, for example limiting redundancy in the training data \cite{lewis2013dramatically} or complementing similarity with diversity \cite{ruder2017learning}.

\paragraph{Sampling sentence pairs}
In the first technique, illustrated in Figure~\ref{fig:dynamic-ds}a, we sample for every epoch $n$ sentence pairs from $G$, using a distribution computed from the domain-specific $\ced_s$ scores. Concretely, this is done as follows: 

First, since higher ranked sentence pairs have lower $\ced_s$ scores, and they can be either negative or positive, we scale and invert $\ced_s$ scores such that $0 \leq \ced'_s \leq1$ for each sentence pair $s\in G$:
\begin{equation}
\small
\label{eq:minmax}
\ced'_s = 1 - 
    \frac{\ced_s - \min(\ced_G)}
           {\max(\ced_G) - \min(\ced_G)},
\end{equation}
where $\ced_G$ refers to the set of $\ced_s$ scores for bitext $G$.

Next, we convert $\ced'_s$ scores to relative weights, such that $\sum_{s\in G} w(s) = 1$:
\begin{equation}
\small
\label{eq:distribution}
w(s) = \frac{\ced'_s}{\sum_{s_i\in G} \ced'_{s_i}}.
\end{equation}
We then use $\{w(s): s\in G\}$ to perform weighted sampling, drawing for each epoch $n$ sentence pairs without replacement. While all selection weights are very close to zero, higher ranked sentences have a noticeably higher probability of being selected than lower-ranked sentences; in practice we find that top-ranked sentences get selected in nearly each epoch, while bottom-ranked sentence pairs get selected at most once.
Note that the sampled selection for any epoch is independent of selections for all other epochs. 

\paragraph{Gradual fine-tuning}
The second dynamic data selection technique, see Figure~\ref{fig:dynamic-ds}b, is inspired by the success of domain-specific fine-tuning \cite{luong2015stanford,zoph2016transfer,sennrich2016improving,freitag2016fast}, in which a model trained on a large general-domain bitext is trained for a few additional epochs only on small in-domain data. However, rather than training a full model on the complete bitext $G$, we gradually decrease the training data size, starting from $G$ and keeping only the top $n$ sentence pairs for the duration of $\eta$ epochs, where the top $n$ pairs are defined by their $\ced_s$ scores. Given its resemblance to fine-tuning, we refer to this variant as \emph{gradual fine-tuning}.

During gradual fine-tuning, the selection size $n$ is a function of epoch $i$:
\begin{equation}
n(i) = \alpha \cdot |G| \cdot \beta^{\lfloor (i-1)/\eta \rfloor}.
\end{equation}
Here $0 \leq \alpha \leq1$ is the \emph{relative start size}, i.e., the fraction of general bitext $G$ used for the first selection, $0 \leq \beta \leq1$ is the \emph{retention rate}, i.e., the fraction of data to be kept in each new selection, and $\eta\geq1$ is the number of consecutive epochs each selected subset is used. Note that $\lfloor i/\eta+1 \rfloor$ indicates rounding down $i/\eta+1$ to the nearest integer. 
For example, if we start with the complete bitext ($\alpha=1$), select the top 60\% ($\beta=0.6$) every second epoch ($\eta=2$), then we run  epochs 1 and 2 with a subset of size $|G|$, epochs 3 and 4 with a subset of size $0.6\cdot|G|$, epochs 5 and 6 with a subset of size $0.36\cdot|G|$, and so on. For every size $n$, the actual selection contains the top $n$ sentences pairs of $G$. 

\section{Experimental settings}
\label{sec:experiments}
We evaluate static and dynamic data selection on a German$\to$English translation task comprising four test sets.
Below we describe the MT systems and data specifications.

\subsection{Machine translation systems}
While the main aim of this paper is to improve data selection for NMT, we also perform comparative experiments using PBMT. Our PBMT system is an in-house system similar to Moses \cite{koehn2007moses}. 
To create optimal PBMT systems given the available resources, we apply test-set-specific parameter tuning using PRO \cite{hopkins2011tuning}. In addition, we use a linearly interpolated target-side language model trained with Kneser-Ney smoothing on 480M tokens of data in various domains. LM interpolation weights are also optimized per test set. 
Consistent with \newcite{axelrod2011domain}, we do not vary the target-side LM between different experiments on the same test set.
All n-gram models in our work are 5-gram.

For our NMT experiments we use an in-house encoder-decoder\footnote{\url{github.com/ketranm/tardis}} 
model with global attention as described in \newcite{luong2015effective}. This choice comes at the cost of optimal translation quality but allows for a relatively fast realization of large-scale experiments given our available resources.
Both the encoder and decoder are four-layer unidirectional LSTMs, with embedding and layer sizes of 1,000. We uniformly initialize all parameters, and use SGD with a mini-batch size of 64 and an initial learning rate of 1, which is decayed by a factor two every epoch after the fifth epoch. We use dropout with probability 0.3, and a beam size of 12. 
We train for 16 epochs and test on the model from the last epoch. All NMT experiments are run on a single NVIDIA Titan X GPU.

\renewcommand{\arraystretch}{1.1}
\begin{table}[h]
\centering
\resizebox{\linewidth}{!}{
\begin{tabular}{@{\,}l @{} c @{} c@{\,\, }c@{} c  @{}c@{\,\, }c@{} c  @{}c@{\,\, }c@{}}
\toprule
&& \multicolumn{2}{c}{Train} && \multicolumn{2}{c}{Dev/valid} && \multicolumn{2}{c}{Test}  \\
\cmidrule{3-4} \cmidrule{6-7} \cmidrule{9-10}
Corpus && Lines & Tokens && Lines & Tokens && Lines & Tokens  \\
\midrule
EMEA &\,\,\,\,\, & 206K & 3.3M &\,\,\,\,\, & 3.9K & 59K &\,\,\,\,\, & 5.8K & 93K  \\
Movies && 101K & 1.2M &\,& 4.5K & 54K && 7.1K & 87K \\
TED && 189K & 3.3M && 2.5K & 50K && 5.4K & 99K  \\
WMT && \,3.8M & \,84M && 3.0K & 64K && 3.0K & 65K  \\
\midrule
Mix && \,4.3M & \,92M && 3.5K & 61K && -- & -- \\
\bottomrule
\end{tabular}}
\caption{Data specifications with tokens counted on the German side. The WMT training corpus contains Commoncrawl, Europarl, and News Commentary but no in-domain news data.} 
\label{table:data}
\end{table}
\renewcommand{\arraystretch}{1}

\begin{figure*}[ht!]
\centering
\includegraphics[width=\textwidth]{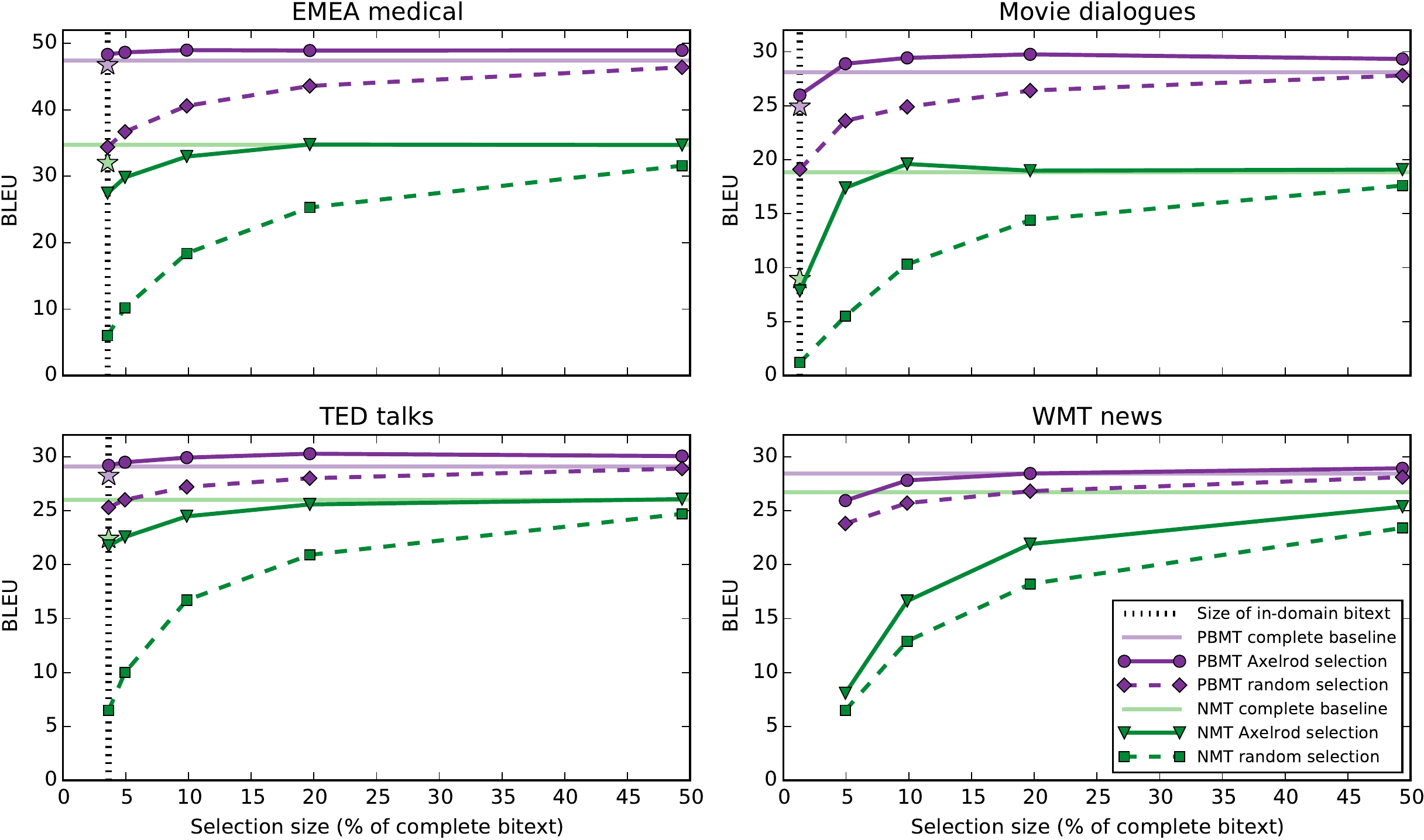}
\caption{PBMT (purple) and NMT (green) German$\to$English results of Axelrod data selection and random data selection (average of three runs) for four domains. Purple and green stars indicate BLEU scores when only the available in-domain data is used. We use selections of the in-domain size $|I|$, and 5\%, 10\%, 20\%, and 50\% of the complete bitext, which are exactly the same for PBMT and NMT.}
\label{fig:axelrod}
\end{figure*}

\subsection{Training and evaluation data}
We evaluate all experiments on four domains: (i) EMEA medical guidelines \cite{tiedemann2009news}, (ii) movie dialogues \cite{vanderwees2016measuring} constructed from OpenSubtitles \cite{lison2016opensubtitles}, (iii) TED talks \cite{cettolo2012web}, and (iv) WMT news. 
For TED, we use IWSLT2010 as development set and IWSLT2011-2014 as test set, and for WMT we use newstest2013 as development set and newstest2016 as test set. 
We train our systems on a mixture of domains, comprising Commoncrawl, Europarl, News Commentary, EMEA, Movies, and TED. Corpus specifications are listed in Table~\ref{table:data}. 

The in-domain LMs used to rank training sentences for data selection are trained on small portions of in-domain parallel data whenever available (3.3M, 1.2M and 3.3M German tokens for EMEA, Movies and TED, respectively).
Since no sizeable in-domain parallel text is available for WMT, we independently sample 200K sentences from the WMT monolingual News Crawl corpora (3.3M German tokens or 3.5M English tokens).
This demonstrates the applicability of data selection techniques even in cases where one lacks parallel in-domain data.

Before running data selection, we preprocess our data by tokenizing, lowercasing and removing sentences that are longer than 50 tokens or that are identified as a different language. After selection, we apply Byte-pair encoding (BPE, \newcite{sennrich2016neural}) with 40K merge operations on either side of the complete mix-of-domains training bitext. For our NMT experiments we use BPE-processed corpora on both bitext sides, while for PBMT we only apply BPE to the German side.
Our NMT systems use a vocabulary size of 40K on both the source and target side. 

\section{Results}
\label{sec:results}
Below we discuss the results of our translation experiments using static and dynamic data selection, measuring translation quality with case-insensitive untokenized BLEU \cite{papineni2002bleu}.

\subsection{Static data selection for PBMT and NMT}
\label{sec:static-results}
We first compare the effects of static data selection with n-gram LMs on both NMT and PBMT using various selection sizes. Concretely, we select the top $n$ sentence pairs such that the number of selected tokens $t\in\left\{5\%, 10\%, 20\%, 50\%\right\}$ of $G$, or $t=|I|$ (the in-domain corpus size).
Figure~\ref{fig:axelrod} shows German$\to$English translation performance in BLEU for our four test sets. 
The benefits of n-gram-based data selection for PBMT (purple circles) are confirmed:
In all test sets, the selection of size $|I|$ (dotted vertical line) yields better performance than using only the in-domain data of the exact same size (purple star), and at least one of the selected subsets---often using only 5\% of the complete bitext---outperforms using the complete bitext (light purple line). We also show that the informed selections are superior to random selections of the same size (purple diamonds).

In NMT, results of n-gram-based data selection (green triangles) vary: While for Movies a selection of only 10\% outperforms the complete bitext (light green line), none of the selected subsets for other test sets is noticeably better than the full bitext.\footnote{Validation cross-entropy converges after 10--12 epochs, never reaching the scores of the complete bitext.}
Interestingly, the same selections of size $|I|$ that proved useful in PBMT, never beat the system that uses exactly the available in-domain data (green star), indicating that the current selections can be further improved for NMT.
In all scenarios we see that NMT suffers much more from small-data settings than PBMT. Finally, the random selections (green squares) show that NMT not only needs large quantities of data, but it is also affected when the selected data is of low quality. In PBMT, both low-quantity and low-quality scenarios appear to be compensated for by the large monolingual LM on the target side.

When comparing the different test sets, we observe that the impact of domain mismatch in NMT with respect to PBMT is largest for the two domains that are most distinct from the general bitext, EMEA and Movies. For WMT, both MT systems achieve very similar baseline results, but translation quality deteriorates considerably in data selection experiments, which is likely caused by the lack of in-domain data in the general bitext.

\paragraph{LSTM versus n-gram}
Before proceeding with dynamic data selection for NMT, we test whether bitext ranking for NMT can be improved using LSTMs rather than conventional n-gram LMs. Table~\ref{table:lstm-vs-ngram} shows NMT BLEU scores of a few different sizes of selected subsets created using n-gram LMs or LSTMs. While results vary among test sets and selection sizes, we observe an average improvement of 0.4 BLEU when using LSTMs instead of n-gram LMs. 
For PBMT, similar results have been reported when replacing n-gram LMs with recurrent neural LMs \cite{duh2013adaptation}.
In all subsequent experiments we use relevance rankings computed with LSTMs instead of n-gram LMs.

\renewcommand{\arraystretch}{1.1}
\begin{table}[h]
\centering
\resizebox{\linewidth}{!}{
\begin{tabular}{l l  cccc}
\toprule
Selection & LM type & EMEA & Movies & TED & WMT \\
\midrule
\multirow{2}{*}{$5\%$} & n-gram & 29.8 & 17.4 & 22.6 & \,\,\,8.1 \\
                                    & LSTM & \textbf{30.0} & \textbf{17.8} & 22.6 & \,\,\,\textbf{9.6} \\
\midrule
\multirow{2}{*}{$10\%$} & n-gram & 33.0 & 19.6 & 24.5 & 16.6 \\
                                      & LSTM & 33.0 & \textbf{19.7} & \textbf{24.7}  & \textbf{17.4} \\
\midrule
\multirow{2}{*}{$20\%$} & n-gram & \textbf{34.8} & 19.0 & 25.6 & 21.9 \\
                                      & LSTM & 34.5 & \textbf{19.6} & \textbf{26.6} & 21.9\\
\bottomrule
\end{tabular}}
\caption{NMT BLEU comparison between using n-gram LMs and LSTMs for bitext ranking. Selection sizes concern the selected bitext subsets; LMs are created from the exact same in-domain data.}
\label{table:lstm-vs-ngram}
\end{table}
\renewcommand{\arraystretch}{1}

\subsection{Dynamic data selection for NMT}
\label{sec:dynamic-results}
Equipped with a relevance ranking of sentence pairs in bitext $G$, we now examine two variants of dynamic data selection as described in Section~\ref{sec:dynamic-method}.

\begin{figure*}[ht!]
\centering
\includegraphics[width=\textwidth]{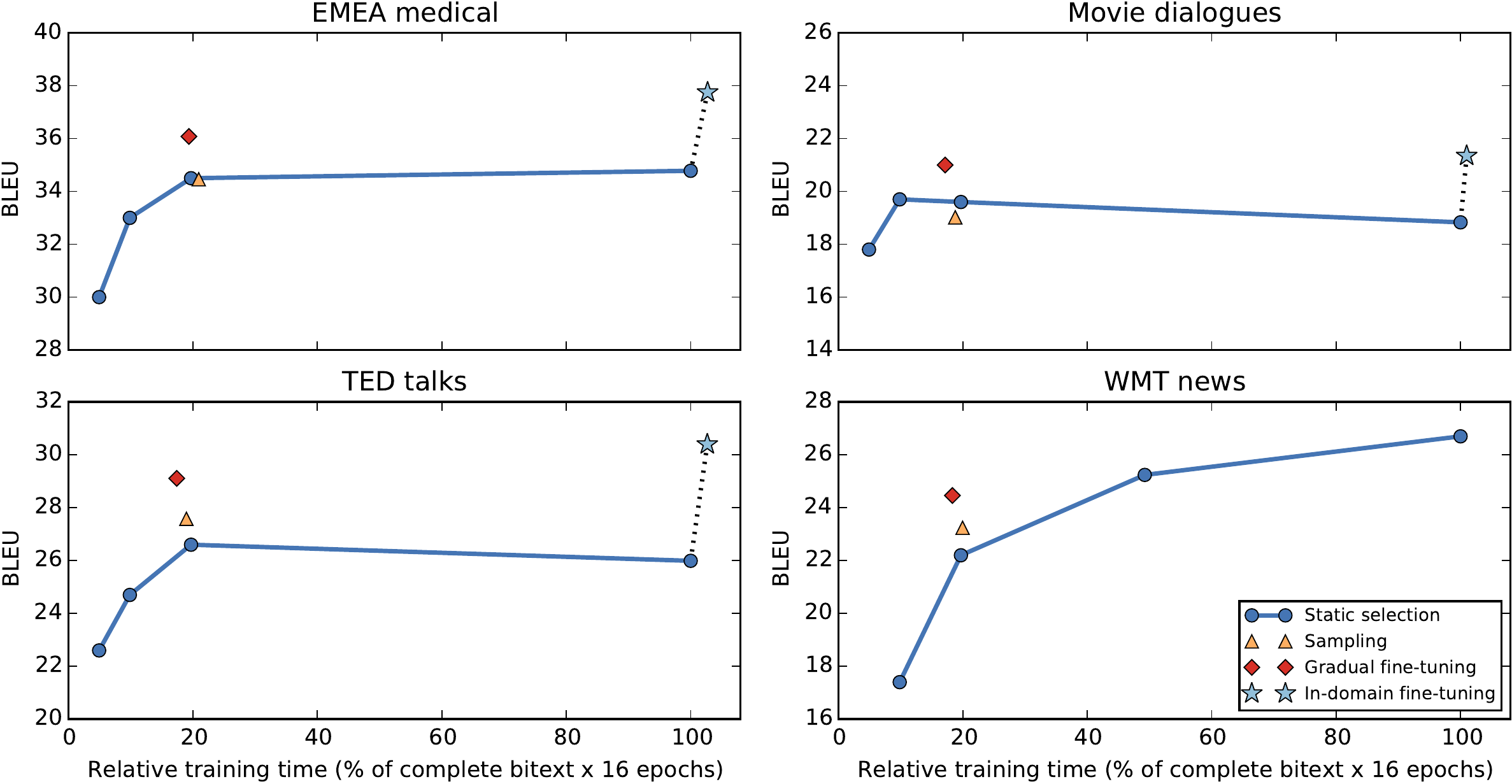}
\caption{Selected German$\to$English translation results of dynamic data selection methods (orange and red markers) compared to conventional static data selection (blue circles). \emph{Relative training time} equals the total number of training tokens relative to the complete baseline, which takes 106 hours to train and is represented by the rightmost blue circle. Note that no parallel in-domain data is available for WMT news. All y-axes are scaled equally for easy comparison of BLEU differences across domains.}
\label{fig:dynamic}
\end{figure*}

\renewcommand{\arraystretch}{1.2}
\begin{table*}[ht!]
\centering
\resizebox{\linewidth}{!}{
\begin{tabular}{@{}c @{\,\, } cc ccc  cccc@{}}
\toprule
\multicolumn{3}{@{}c}{Experiment} & & \raisebox{-0.6ex}{Relative} && \multicolumn{4}{c@{}}{BLEU} \\
\cmidrule{1-3} \cmidrule{7-10}
Start size & Retention rate $\beta$ & Decrease every && \raisebox{0.7ex}{training time} && EMEA & Movies & TED & WMT \\
\midrule
\multicolumn{3}{@{}l}{\emph{Static selection top 20\%}} && 20\% && 34.5 & 19.6 & 26.6 & 21.9 \\
\midrule
50\% ($\alpha=0.5$) & 0.7 & $\eta=2$ epochs && 18--20\%       && \textbf{36.1 (+1.6)} & 21.0 (+1.4) & \textbf{29.1 (+2.5)} & 24.5 (+2.6) \\
50\% ($\alpha=0.5$) & 0.5 & $\eta=4$ epochs && 21--23\%       && 36.0 (+1.5) & \textbf{21.2 (+1.6)} & 29.0 (+2.4) & 25.0 (+3.1) \\
50\% ($\alpha=0.5$) & 0.6 & $\eta=4$ epochs && 25--27\%       && 35.6 (+1.1) & 21.0 (+1.4) & 28.5 (+1.9) & 25.1 (+3.2) \\
\midrule
100\% \,\,($\alpha=1$) & 0.6 & $\eta=2$ epochs && 29--31\%     && 35.5 (+1.0) & 21.1 (+1.5) & 29.0 (+2.4) & 25.6 (+3.7) \\
100\% \,\,($\alpha=1$) & 0.7 & $\eta=2$ epochs && 37--39\%     && 35.9 (+1.4) &  20.4 (+0.8) & 28.2 (+1.6) & 25.8 (+3.9) \\
100\% \,\,($\alpha=1$) & 0.9 & $\eta=1$ epoch\,\, &&  50--52\% && 35.4 (+0.9) & 19.6 ($\pm$0.0) & 27.4 (+0.8) & \textbf{26.1 (+4.2)} \\
\midrule
\multicolumn{3}{@{}l}{\emph{Complete bitext baseline}} && 100\% && 34.8 & 18.8 & 26.0 & 26.7 \\
\multicolumn{3}{@{}l}{\emph{Gold: fine-tuning on in-domain data}} && 101--103\% && 37.7 & 21.3 & 30.4 & -- \\
\bottomrule
\end{tabular}}
\caption{German$\to$English BLEU results of various gradual fine-tuning experiments sorted by relative training time. Indicated improvements are with respect to static selection using 20\% of the bitext, and highest scores per test set are bold-faced. 
Results from the first experiment are also shown in Figure~\ref{fig:dynamic}.}
\label{table:gradual}
\end{table*}
\renewcommand{\arraystretch}{1}

We are interested in reducing training time while limiting the negative effect on BLEU for various domains. Therefore we report BLEU as well as the \emph{relative training time} of each experiment.
Since wall-clock times depend on other factors such as the NMT architecture and memory speed, we define training time as the total number of tokens observed while training the NMT system, i.e., the sum of tokens in the selected subsets of all epochs. 
We report all training times relative to the training time of our complete-bitext baseline (i.e., 4.3M tokens $\times$ 16 epochs).
Note that this measure of training time corresponds closely but not exactly to the number of model updates, as the latter relies on the number of sentences, which vary in length, rather than the number of tokens in the training data. 
For completeness: Training the 100\% baseline takes 106 hours, while our fastest dynamic selection variant takes 19--21 hours. Computing CED scores takes $\sim$15 minutes when using n-gram LMs and 5--6 hours when using LSTMs.

Figure~\ref{fig:dynamic} shows BLEU scores of some selected experiments as a function of relative training time. Compared to static data selection (blue lines), our weighted sampling technique (orange triangles) yields variable results. When sampling a subset of 20\% of $|G|$ from the top 50\% of the ranked bitext, we obtain small improvements for TED and WMT, but small drops for EMEA and Movies. Other selection sizes (30\% and 40\%, not shown) give similar results lacking a consistent pattern.

By contrast, our gradual fine-tuning method performs consistently better than static selection, and even beats the general baseline in three out of four test sets. The displayed version uses settings $(\alpha=0.5, \beta=0.7, \eta=2$) and is at least as fast as static selection using 20\% of the bitext, yielding up to +2.6 BLEU improvement (for WMT news) over this static version. Compared to the complete baseline, this gradual fine-tuning method improves up to +3.1 BLEU (for TED talks).

Table~\ref{table:gradual} provides detailed information on additional experiments using other settings. For all three test domains which are covered in the parallel data---EMEA, Movies and TED---improvements are highest when starting gradual fine-tuning with only the top 50\% of the ranked bitext, which are also the fastest approaches.
For WMT, which is not covered in the general bitext, adding more data clearly benefits translation quality. 
These findings are consistent with the static data selection patterns; Using low-ranked sentences on top of the most relevant selection does not improve translation performance for any domain except WMT news.

Finally, we compare our data selection experiments to domain-specific fine-tuning (light blue stars in Figure~\ref{fig:dynamic}), which is the current state-of-the-art for domain adaptation in NMT. To this end, we first train a model on the complete bitext, and then train for twelve additional epochs on available in-domain data, using an initial learning rate of 1 which halves every epoch.
Depending on the test set, this approach yields +2.5--4.4 BLEU improvements over our baselines, however it does not speed up training and requires a parallel in-domain text which may not be available (e.g., for WMT). While none of our data selection experiments outperforms domain-specific fine-tuning, we obtain competitive translation quality in only 20\% of the training time.
In additional experiments we found that in-domain fine-tuning on top of our selection approaches does not yield improvements.

\section{Further analysis}
\label{sec:analysis}
In this section we conduct a few additional experiments and analyses.
We restrict to one parameter setting per selection approach: Static selection and sampling with 20\% of the data, and gradual fine-tuning using $(\alpha=0.5, \beta=0.7, \eta=2$). All have very similar training times.

\begin{figure}[h]
\centering
\includegraphics[width=\linewidth]{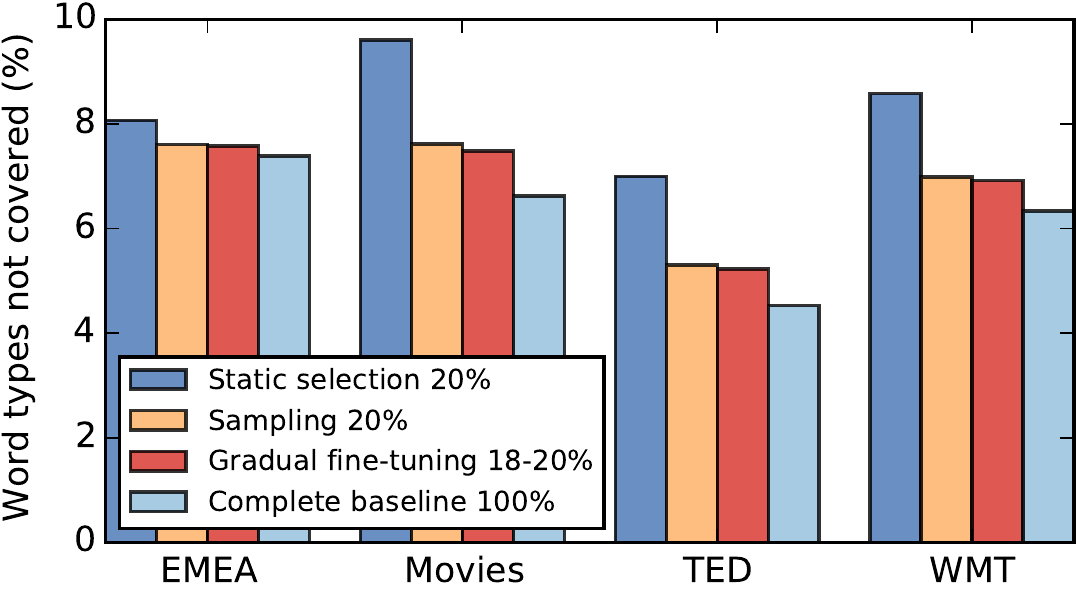}
\caption{Test set source words not covered in the training data of different data selection methods.}
\label{fig:coverage}
\end{figure}

First, we hypothesize that dynamic data selection works well because more different sentence pairs are observed during training, and it therefore increases coverage with respect to static data selection. To verify this, we measure for each test set the number of unseen source word types in the training data of different selection methods.
Figure~\ref{fig:coverage} shows indeed that the average number of unseen word types is reduced noticeably in both of our dynamic selection techniques, being much closer to the complete bitext baseline than to static selection. Note that all methods use the same  vocabulary during training.

Next, following the static data selection experiments in Section~\ref{sec:static-results}, we examine how well dynamic data selection  performs using random selections. To this end, we repeat all techniques using a bitext which is ranked randomly rather than by its relevance to the test sets. The results in Table~\ref{table:analysis} show that the bitext ranking plays a crucial role in the success of data selection. However, the results also show that \emph{even} in the absence of an appropriate bitext ranking, dynamic data selection---and in particular gradual fine-tuning---is still superior to static data selection. 
We explain this result as follows: Compared to
static selection, both sampling and gradual fine-tuning have better coverage 
due to their improved exploration of the data. However, sampling also
suffers from a surprise effect of observing new data in every
epoch. Gradual fine-tuning on the other hand gradually improves learning on a
subset of the selected data, suggesting that repetition across epochs
has a positive effect on translation quality.

\renewcommand{\arraystretch}{1.1}
\begin{table}[h]
\centering
\resizebox{\linewidth}{!}{
\begin{tabular}{@{\,} l @{\,\,\,\, } l @{\,\,\,\, } c@{\,\,\, }c@{\,\,\, }c@{\,\,\, }c @{\,}}
\toprule
Ranking & Method & EMEA & Movies & TED & WMT \\
\midrule
\multirow{3}{*}{Relevance} & Gradual FT       & \textbf{36.1} & \textbf{21.0} & \textbf{29.1} & \textbf{24.5} \\
                                               & Sampling 20\%  & 34.5 & 19.0 & 27.6 & 23.2 \\
                                               & Static 20\%        & 34.5 & 19.6 & 26.6 & 21.9 \\
\midrule
\multirow{3}{*}{Random}        & Gradual FT       & \textbf{29.2} & \textbf{16.1} & \textbf{23.2} & \textbf{21.3} \\
                                               & Sampling 20\%  & 26.7 & 14.4 & 22.0 & 19.8 \\
                                               & Static 20\%        & 25.3 & 14.4 & 20.9 & 18.2 \\
\bottomrule
\end{tabular}}
\caption{BLEU scores of data selection using relevance versus random ranking of the bitext. Gradual fine-tuning uses $(\alpha=0.5, \beta=0.7, \eta=2$), with relative training times of 18--20\%.}
\label{table:analysis}
\end{table}
\renewcommand{\arraystretch}{1}

\begin{figure}[h]
\centering
\includegraphics[width=\linewidth]{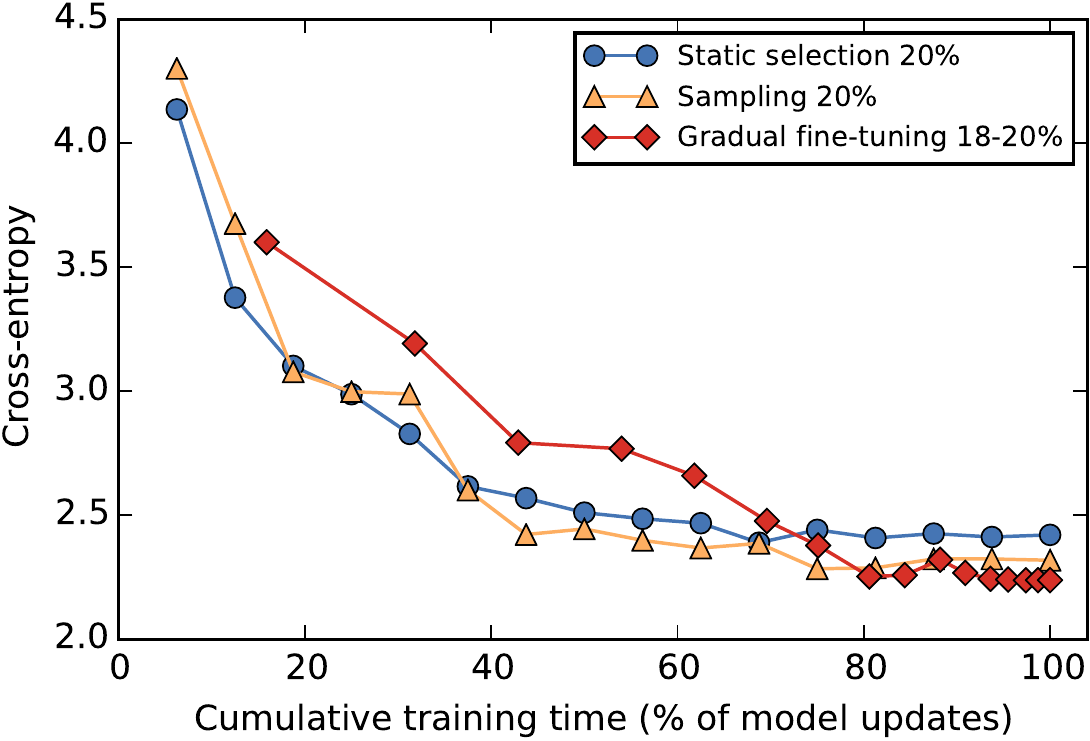}
\caption{German$\to$English cross-entropy of the TED dev set as a function of training time. Each data point represents a completed training epoch.}
\label{fig:ppl}
\end{figure}

One could expect that changing the data during training results in volatile training behavior. To test this, we inspect cross-entropy of our development sets after every training epoch. 
Figure~\ref{fig:ppl} shows these results for TED. Clearly, static data selection converges most steadily. However, both dynamic selection techniques eventually converge to a lower cross-entropy value which is reflected by higher translation quality of the test set. We observe very similar behavior for the other test sets.

By its nature, our gradual fine-tuning technique uses training epochs of different sizes, and therefore also implicitly differs from other methods in its parameter optimization behavior. 
Since we decrease both the training data size and the SGD learning rate after finishing complete training epochs, we automatically decay the learning rate at decreasing time intervals. 
We therefore study how this approach is affected when we (i) decay the learning rate after a fixed number of updates (i.e., the same as in static data selection) rather than per epoch, or (ii) keep the learning rate fixed. 
In the first scenario, we observe that translation performance drops with --1.1--2.0 BLEU. When keeping a fixed learning rate, BLEU scores hardly change or even improve, indicating that the implicit change in search behavior may contribute to the success of gradual fine-tuning. 

\section{Related work}
\label{sec:rw}
A few research topics are related to our work. Regarding data selection for SMT, previous work has targeted two goals; to reduce model sizes and training times, or to adapt to new domains. Data selection methods for domain adaptation mostly employ information theory metrics to rank training sentences by their relevance to the domain at hand. This has been applied monolingually \cite{gao2002toward} as well as bilingually \cite{yasuda2008method}. 
In more recent work, training sentences are typically ranked according to their cross-entropy \emph{difference} between in-domain and general-domain data \cite{moore2010intelligent, axelrod2011domain,axelrod2015data}, favoring sentences that are similar to the test domain and at the same time dissimilar from the general domain. \newcite{duh2013adaptation} and \newcite{chen2016semi} present similar methods in which n-gram LMs are replaced by neural LMs or neural classifiers, respectively.

Data selection with the aim of model size and training time reduction has the objective to use the minimum amount of data while still maintaining high vocabulary coverage \cite{eck2005low,gasco2012does,lewis2013dramatically}. 
In a comparative study, \newcite{mirkin2014data} find that similarity-objected methods perform best if the test domain and general corpus are very different, while a coverage-objected method is superior if test and general corpus are relatively similar.
A comprehensive survey on data selection for SMT is provided by \newcite{eetemadi2015survey}.
While in this work we have used a similarity objective to rank our bitext, one could also apply dynamic data selection using a coverage objective.

In NMT, data selection can serve similar goals as in PBMT; increasing training efficiency or domain adaptation. 
Domain adaptation in NMT typically involves training a model on the complete bitext, followed by fine-tuning the parameters on a smaller in-domain corpus \cite{luong2015stanford,zoph2016transfer}.
Other work combines fine-tuning with model ensembles \cite{freitag2016fast} or with domain-specific tags in the training corpus \cite{chu2017empirical}. Finally, \newcite{sennrich2016improving} adapt their systems by back-translating in-domain data, which is then added to the training data and used for fine-tuning.

Some other previous work has addressed training efficiency for NMT, for example by parallelizing models or data \cite{wu2016google}, modifying the NMT network structure \cite{kalchbrenner2016neural}, decreasing the number of parameters through knowledge distillation \cite{crego2016systran,kim2016sequence}, or by boosting parts of the data that are `challenging' to the NMT system \cite{zhang2016boosting}. The latter is most related to our work since training data is also adjusted during training, however we reduce the training data size much more aggressively and study different techniques of data selection. 

Finally, recent work comparing various aspects for PBMT and NMT includes \cite{bentivogli2016neural,farajian2017neural,toral2017multi,koehn2017six}.

\section{Conclusions}
\label{sec:conclusions}
With the recent increase in popularity of neural machine translation (NMT), we explored in this paper \emph{to what extent} and \emph{how} NMT can benefit from data selection.
We first showed that a state-of-the-art data selection method yields unreliable results for NMT while consistently performing well for PBMT. Next, we have introduced \emph{dynamic data selection} for NMT, which entails varying the selected subset of training data between different training epochs. 
We explored two techniques of dynamic data selection and found that our \emph{gradual fine-tuning} technique, in which we gradually reduce training size, improves consistently over conventional static data selection (up to +2.6 BLEU) and over a high-resource general baseline (up to +3.1 BLEU).
Moreover, gradual fine-tuning approximates in-domain fine-tuning using only $\sim$20\% of the training time, even when no parallel in-domain data is available.

\section*{Acknowledgments}
This research was funded in part by NWO under project numbers 639.022.213 and 639.021.646.
We thank Ke Tran for providing the NMT system, and the reviewers for their valuable comments.

\bibliography{emnlp2017}

\begin{thebibliography}{42}
\expandafter\ifx\csname natexlab\endcsname\relax\def\natexlab#1{#1}\fi

\bibitem[{Axelrod et~al.(2011)Axelrod, He, and Gao}]{axelrod2011domain}
Amittai Axelrod, Xiaodong He, and Jianfeng Gao. 2011.
\newblock Domain adaptation via pseudo in-domain data selection.
\newblock In \emph{Proceedings of the 2011 Conference on Empirical Methods in
  Natural Language Processing}, pages 355--362.

\bibitem[{Axelrod et~al.(2015)Axelrod, Resnik, He, and
  Ostendorf}]{axelrod2015data}
Amittai Axelrod, Philip Resnik, Xiaodong He, and Mari Ostendorf. 2015.
\newblock Data selection with fewer words.
\newblock In \emph{Proceedings of the Tenth Workshop on Statistical Machine
  Translation}, pages 58--65.

\bibitem[{Bahdanau et~al.(2014)Bahdanau, Cho, and Bengio}]{bahdanau2014neural}
Dzmitry Bahdanau, Kyunghyun Cho, and Yoshua Bengio. 2014.
\newblock Neural machine translation by jointly learning to align and
  translate.
\newblock \emph{arXiv preprint arXiv:1409.0473}.

\bibitem[{Bentivogli et~al.(2016)Bentivogli, Bisazza, Cettolo, and
  Federico}]{bentivogli2016neural}
Luisa Bentivogli, Arianna Bisazza, Mauro Cettolo, and Marcello Federico. 2016.
\newblock Neural versus phrase-based machine translation quality: a case study.
\newblock In \emph{Proceedings of the 2016 Conference on Empirical Methods in
  Natural Language Processing}, pages 257--267.

\bibitem[{Bojar et~al.(2016)Bojar, Chatterjee, Federmann, Graham, Haddow, Huck,
  Yepes, Koehn, Logacheva, Monz et~al.}]{bojar2016findings}
Ondrej Bojar, Rajen Chatterjee, Christian Federmann, Yvette Graham, Barry
  Haddow, Matthias Huck, Antonio~Jimeno Yepes, Philipp Koehn, Varvara
  Logacheva, Christof Monz, et~al. 2016.
\newblock Findings of the 2016 conference on machine translation ({WMT}16).
\newblock In \emph{Proceedings of the first conference on machine translation
  (WMT16)}.

\bibitem[{Cettolo et~al.(2012)Cettolo, Girardi, and Federico}]{cettolo2012web}
Mauro Cettolo, Christian Girardi, and Marcello Federico. 2012.
\newblock Wit$^3$: Web inventory of transcribed and translated talks.
\newblock In \emph{Proceedings of the 16$^{th}$ Conference of the European
  Association for Machine Translation (EAMT)}, pages 261--268.

\bibitem[{Chen and Huang(2016)}]{chen2016semi}
Boxing Chen and Fei Huang. 2016.
\newblock Semi-supervised convolutional networks for translation adaptation
  with tiny amount of in-domain data.
\newblock In \emph{Proceedings of the 20th SIGNLL Conference on Computational
  Natural Language Learning (CoNLL)}, pages 314--323.

\bibitem[{Cho et~al.(2014)Cho, van Merrienboer, Gulcehre, Bahdanau, Bougares,
  Schwenk, and Bengio}]{cho2014learning}
Kyunghyun Cho, Bart van Merrienboer, Caglar Gulcehre, Dzmitry Bahdanau, Fethi
  Bougares, Holger Schwenk, and Yoshua Bengio. 2014.
\newblock Learning phrase representations using {RNN} encoder--decoder for
  statistical machine translation.
\newblock In \emph{Proceedings of the 2014 Conference on Empirical Methods in
  Natural Language Processing (EMNLP)}, pages 1724--1734.

\bibitem[{Chu et~al.(2017)Chu, Dabre, and Kurohashi}]{chu2017empirical}
Chenhui Chu, Raj Dabre, and Sadao Kurohashi. 2017.
\newblock An empirical comparison of simple domain adaptation methods for
  neural machine translation.
\newblock In \emph{Proceedings of the 55th Annual Meeting of the Association
  for Computational Linguistics}.

\bibitem[{Crego et~al.(2016)Crego, Kim, Klein, Rebollo, Yang, Senellart,
  Akhanov, Brunelle, Coquard, Deng et~al.}]{crego2016systran}
Josep Crego, Jungi Kim, Guillaume Klein, Anabel Rebollo, Kathy Yang, Jean
  Senellart, Egor Akhanov, Patrice Brunelle, Aurelien Coquard, Yongchao Deng,
  et~al. 2016.
\newblock {SYSTRAN}'s pure neural machine translation systems.
\newblock \emph{arXiv preprint arXiv:1610.05540}.

\bibitem[{Duh et~al.(2013)Duh, Neubig, Sudoh, and Tsukada}]{duh2013adaptation}
Kevin Duh, Graham Neubig, Katsuhito Sudoh, and Hajime Tsukada. 2013.
\newblock Adaptation data selection using neural language models: Experiments
  in machine translation.
\newblock In \emph{Proceedings of the 51st Annual Meeting of the Association
  for Computational Linguistics}, pages 678--683.

\bibitem[{Eck et~al.(2005)Eck, Vogel, and Waibel}]{eck2005low}
Matthias Eck, Stephan Vogel, and Alex Waibel. 2005.
\newblock Low cost portability for statistical machine translation based on
  n-gram frequency and {TF-IDF}.
\newblock In \emph{Proceedings of the 2005 International Workshop on Spoken
  Language Translation}, pages 61--67.

\bibitem[{Eetemadi et~al.(2015)Eetemadi, Lewis, Toutanova, and
  Radha}]{eetemadi2015survey}
Sauleh Eetemadi, William Lewis, Kristina Toutanova, and Hayder Radha. 2015.
\newblock Survey of data-selection methods in statistical machine translation.
\newblock \emph{Machine Translation}, 29(3-4):189--223.

\bibitem[{Fadaee et~al.(2017)Fadaee, Bisazza, and Monz}]{fadaee2017data}
Marzieh Fadaee, Arianna Bisazza, and Christof Monz. 2017.
\newblock Data augmentation for low-resource neural machine translation.
\newblock In \emph{Proceedings of the 55th Annual Meeting of the Association
  for Computational Linguistics}.

\bibitem[{Farajian et~al.(2017)Farajian, Turchi, Negri, Bertoldi, and
  Federico}]{farajian2017neural}
M.~Amin Farajian, Marco Turchi, Matteo Negri, Nicola Bertoldi, and Marcello
  Federico. 2017.
\newblock Neural vs. phrase-based machine translation in a multi-domain
  scenario.
\newblock In \emph{Proceedings of the 15th Conference of the European Chapter
  of the Association for Computational Linguistics}, pages 280--284.

\bibitem[{Freitag and Al-Onaizan(2016)}]{freitag2016fast}
Markus Freitag and Yaser Al-Onaizan. 2016.
\newblock Fast domain adaptation for neural machine translation.
\newblock \emph{arXiv preprint arXiv:1612.06897}.

\bibitem[{Gao et~al.(2002)Gao, Goodman, Li, and Lee}]{gao2002toward}
Jianfeng Gao, Joshua Goodman, Mingjing Li, and Kai-Fu Lee. 2002.
\newblock Toward a unified approach to statistical language modeling for
  chinese.
\newblock \emph{ACM Transactions on Asian Language Information Processing
  (TALIP)}, 1(1):3--33.

\bibitem[{Gasc\'{o} et~al.(2012)Gasc\'{o}, Rocha, Sanchis-Trilles,
  Andr\'{e}s-Ferrer, and Casacuberta}]{gasco2012does}
Guillem Gasc\'{o}, Martha-Alicia Rocha, Germ\'{a}n Sanchis-Trilles, Jes\'{u}s
  Andr\'{e}s-Ferrer, and Francisco Casacuberta. 2012.
\newblock Does more data always yield better translations?
\newblock In \emph{Proceedings of the 13th Conference of the European Chapter
  of the Association for Computational Linguistics}, pages 152--161.

\bibitem[{Hopkins and May(2011)}]{hopkins2011tuning}
Mark Hopkins and Jonathan May. 2011.
\newblock Tuning as ranking.
\newblock In \emph{Proceedings of the 2011 Conference on Empirical Methods in
  Natural Language Processing}, pages 1352--1362.

\bibitem[{Kalchbrenner et~al.(2016)Kalchbrenner, Espeholt, Simonyan, Oord,
  Graves, and Kavukcuoglu}]{kalchbrenner2016neural}
Nal Kalchbrenner, Lasse Espeholt, Karen Simonyan, Aaron van~den Oord, Alex
  Graves, and Koray Kavukcuoglu. 2016.
\newblock Neural machine translation in linear time.
\newblock \emph{arXiv preprint arXiv:1610.10099}.

\bibitem[{Kim and Rush(2016)}]{kim2016sequence}
Yoon Kim and Alexander~M. Rush. 2016.
\newblock Sequence-level knowledge distillation.
\newblock In \emph{Proceedings of the 2016 Conference on Empirical Methods in
  Natural Language Processing}, pages 1317--1327.

\bibitem[{Koehn et~al.(2007)Koehn, Hoang, Birch, Callison-Burch, Federico,
  Bertoldi, Cowan, Shen, Moran, Zens, Dyer, Bojar, Constantin, and
  Herbst}]{koehn2007moses}
Philipp Koehn, Hieu Hoang, Alexandra Birch, Chris Callison-Burch, Marcello
  Federico, Nicola Bertoldi, Brooke Cowan, Wade Shen, Christine Moran, Richard
  Zens, Chris Dyer, Ondrej Bojar, Alexandra Constantin, and Evan Herbst. 2007.
\newblock Moses: Open source toolkit for statistical machine translation.
\newblock In \emph{Proceedings of the 45th Annual Meeting of the Association
  for Computational Linguistics Demo and Poster Sessions}, pages 177--180.

\bibitem[{Koehn and Knowles(2017)}]{koehn2017six}
Philipp Koehn and Rebecca Knowles. 2017.
\newblock Six challenges for neural machine translation.
\newblock \emph{arXiv preprint arXiv:1706.03872}.

\bibitem[{Lewis and Eetemadi(2013)}]{lewis2013dramatically}
William~D. Lewis and Sauleh Eetemadi. 2013.
\newblock Dramatically reducing training data size through vocabulary
  saturation.
\newblock In \emph{Proceedings of the 8th Workshop on Statistical Machine
  Translation}, pages 281--291.

\bibitem[{Lison and Tiedemann(2016)}]{lison2016opensubtitles}
Pierre Lison and J\"{o}rg Tiedemann. 2016.
\newblock Opensubtitles2016: Extracting large parallel corpora from movie and
  tv subtitles.
\newblock In \emph{Proceedings of the 10th International Conference on Language
  Resources and Evaluation (LREC 2016)}.

\bibitem[{Luong and Manning(2015)}]{luong2015stanford}
Minh-Thang Luong and Christopher~D Manning. 2015.
\newblock Stanford neural machine translation systems for spoken language
  domains.
\newblock In \emph{Proceedings of the 12th International Workshop on Spoken
  Language Translation}, pages 76--79.

\bibitem[{Luong et~al.(2015{\natexlab{a}})Luong, Pham, and
  Manning}]{luong2015effective}
Minh-Thang Luong, Hieu Pham, and Christopher~D. Manning. 2015{\natexlab{a}}.
\newblock Effective approaches to attention-based neural machine translation.
\newblock In \emph{Proceedings of the 2015 Conference on Empirical Methods in
  Natural Language Processing}, pages 1412--1421.

\bibitem[{Luong et~al.(2015{\natexlab{b}})Luong, Sutskever, Le, Vinyals, and
  Zaremba}]{luong2015addressing}
Minh-Thang Luong, Ilya Sutskever, Quoc Le, Oriol Vinyals, and Wojciech Zaremba.
  2015{\natexlab{b}}.
\newblock Addressing the rare word problem in neural machine translation.
\newblock In \emph{Proceedings of the 53rd Annual Meeting of the Association
  for Computational Linguistics and the 7th International Joint Conference on
  Natural Language Processing (Volume 1: Long Papers)}, pages 11--19.

\bibitem[{Mirkin and Besacier(2014)}]{mirkin2014data}
Shachar Mirkin and Laurent Besacier. 2014.
\newblock Data selection for compact adapted {SMT} models.
\newblock In \emph{Proceedings of the 11th Conference of the Association for
  Machine Translation in the Americas}, pages 301--314.

\bibitem[{Moore and Lewis(2010)}]{moore2010intelligent}
Robert~C. Moore and William Lewis. 2010.
\newblock Intelligent selection of language model training data.
\newblock In \emph{Proceedings of the ACL 2010 Conference Short Papers}, pages
  220--224.

\bibitem[{Papineni et~al.(2002)Papineni, Roukos, Ward, and
  Zhu}]{papineni2002bleu}
Kishore Papineni, Salim Roukos, Todd Ward, and Wei-Jing Zhu. 2002.
\newblock Bleu: a method for automatic evaluation of machine translation.
\newblock In \emph{Proceedings of 40th Annual Meeting of the Association for
  Computational Linguistics}, pages 311--318.

\bibitem[{Ruder and Plank(2017)}]{ruder2017learning}
Sebastian Ruder and Barbara Plank. 2017.
\newblock Learning to select data for transfer learning with bayesian
  optimization.
\newblock In \emph{Proceedings of the 2017 Conference on Empirical Methods in
  Natural Language Processing}.

\bibitem[{Sennrich et~al.(2016{\natexlab{a}})Sennrich, Haddow, and
  Birch}]{sennrich2016improving}
Rico Sennrich, Barry Haddow, and Alexandra Birch. 2016{\natexlab{a}}.
\newblock Improving neural machine translation models with monolingual data.
\newblock In \emph{Proceedings of the 54th Annual Meeting of the Association
  for Computational Linguistics}, pages 86--96.

\bibitem[{Sennrich et~al.(2016{\natexlab{b}})Sennrich, Haddow, and
  Birch}]{sennrich2016neural}
Rico Sennrich, Barry Haddow, and Alexandra Birch. 2016{\natexlab{b}}.
\newblock Neural machine translation of rare words with subword units.
\newblock In \emph{Proceedings of the 54th Annual Meeting of the Association
  for Computational Linguistics}, pages 1715--1725.

\bibitem[{Sutskever et~al.(2014)Sutskever, Vinyals, and
  Le}]{sutskever2014sequence}
Ilya Sutskever, Oriol Vinyals, and Quoc~V. Le. 2014.
\newblock Sequence to sequence learning with neural networks.
\newblock In \emph{Advances in neural information processing systems}, pages
  3104--3112.

\bibitem[{Tiedemann(2009)}]{tiedemann2009news}
J{\"o}rg Tiedemann. 2009.
\newblock News from {OPUS}-a collection of multilingual parallel corpora with
  tools and interfaces.
\newblock In \emph{Recent advances in natural language processing}, volume~5,
  pages 237--248.

\bibitem[{Toral and S\'{a}nchez-Cartagena(2017)}]{toral2017multi}
Antonio Toral and V\'{i}ctor~M. S\'{a}nchez-Cartagena. 2017.
\newblock A multifaceted evaluation of neural versus phrase-based machine
  translation for 9 language directions.
\newblock In \emph{Proceedings of the 15th Conference of the European Chapter
  of the Association for Computational Linguistics: Volume 1, Long Papers},
  pages 1063--1073.

\bibitem[{van~der Wees et~al.(2016)van~der Wees, Bisazza, and
  Monz}]{vanderwees2016measuring}
Marlies van~der Wees, Arianna Bisazza, and Christof Monz. 2016.
\newblock Measuring the effect of conversational aspects on machine translation
  quality.
\newblock In \emph{Proceedings of COLING 2016, the 26th International
  Conference on Computational Linguistics}, pages 2571--2581.

\bibitem[{Wu et~al.(2016)Wu, Schuster, Chen, Le, Norouzi, Macherey, Krikun,
  Cao, Gao, Macherey et~al.}]{wu2016google}
Yonghui Wu, Mike Schuster, Zhifeng Chen, Quoc~V Le, Mohammad Norouzi, Wolfgang
  Macherey, Maxim Krikun, Yuan Cao, Qin Gao, Klaus Macherey, et~al. 2016.
\newblock Google's neural machine translation system: {B}ridging the gap
  between human and machine translation.
\newblock \emph{arXiv preprint arXiv:1609.08144}.

\bibitem[{Yasuda et~al.(2008)Yasuda, Zhang, Yamamoto, and
  Sumit}]{yasuda2008method}
Keiji Yasuda, Ruiqiang Zhang, Hirofumi Yamamoto, and Eiichiro Sumit. 2008.
\newblock Method of selecting training data to build a compact and efficient
  translation model.
\newblock In \emph{International Joint Conference on Natural Language
  Processing (IJCNLP)}, pages 655--660.

\bibitem[{Zhang et~al.(2016)Zhang, Kim, Crego, and
  Senellart}]{zhang2016boosting}
Dakun Zhang, Jungi Kim, Joseph Crego, and Jean Senellart. 2016.
\newblock Boosting neural machine translation.
\newblock \emph{arXiv preprint arXiv:1612.06138}.

\bibitem[{Zoph et~al.(2016)Zoph, Yuret, May, and Knight}]{zoph2016transfer}
Barret Zoph, Deniz Yuret, Jonathan May, and Kevin Knight. 2016.
\newblock Transfer learning for low-resource neural machine translation.
\newblock \emph{arXiv preprint arXiv:1604.02201}.

\end{thebibliography}
\bibliographystyle{emnlp_natbib}

\end{document}